# FMT: A Multimodal Pneumonia Detection Model Based on Stacking MOE Framework

Jingyu Xu*
School of Informatics, Computing and Cyber Systems, Northern Arizona University, Arizona, U.S
jyxu01@outlook.com

Yang Wang
Cluster BI Inc, Toronto, Canada
wangyrick100@gmail.com

*Abstract*—Artificial intelligence has shown the potential to improve diagnostic accuracy through medical image analysis for pneumonia diagnosis. However, traditional multimodal approaches often fail to address real-world challenges such as incomplete data and modality loss. In this study, a Flexible Multimodal Transformer (FMT) was proposed, which uses ResNet-50 and BERT for joint representation learning, followed by a dynamic masked attention strategy that simulates clinical modality loss to improve robustness; finally, a sequential mixture of experts (MOE) architecture was used to achieve multi-level decision refinement. After evaluation on a small multimodal pneumonia dataset, FMT achieved state-of-the-art performance with 94% accuracy, 95% recall, and 93% F1 score, outperforming single-modal baselines (ResNet: 89%; BERT: 79%) and the medical benchmark CheXMed (90%), providing a scalable solution for multimodal diagnosis of pneumonia in resource-constrained medical settings.

## I. INTRODUCTION

Pneumonia is a global disease caused by various microorganisms that leads to inflammation of the lung parenchyma[1]. Early detection and diagnosis are crucial for effective treatment and prognosis, as pneumonia can severely affect patients' quality of life [2]. Some common diagnostic methods for pneumonia include chest X-rays, chest computed tomography (CT), and chest magnetic resonance imaging (MRI). Although chest X-rays have lower sensitivity in detecting pneumonia compared to chest CT and MRI, as they only provide planar imaging, chest X-rays are still the most widely used diagnostic tool in clinical practice worldwide due to their cost-effectiveness [3].

The introduction of artificial intelligence technology has greatly promoted the diversified data analysis in various fields[4-6]. With the advancement of deep learning technology, convolutional neural networks (CNNs) have become widely used in medical imaging, particularly for pneumonia detection. CNN is a deep learning model whose core components are convolution and pooling, and is good at image extraction and classification [7]. In CNN, the convolution kernel is multiplied with the local area of the input data step by step and the sum is generated. The stride is used to control the movement speed of the convolution kernel in the process.

With the development of large models, artificial intelligence has brought unstructured data diagnosis to pneumonia diagnosis, especially on images. Image analysis using CNNs has made significant progress in improving the accuracy of pneumonia diagnosis and assessment on chest radiographs. This progress is mainly attributed to the evolution of deep learning, including the use of pre-trained CNN architectures for transfer learning and the integration of multiple components for better feature extraction [8]. However, One of the challenges in using CNN for pneumonia diagnosis is the sparsity of relevant medical data [9]. In recent years, the development of multimodal diagnostic methods has made significant progress in medicine. These methods overcome the limitations of insufficient single-modal data by combining multiple clinical data types, thereby enabling a more complete and comprehensive assessment of a patient's condition.

In this paper, we introduce the multi-feature transformer model called Flexible Multimodal Transformer (FMT), which is innovatively designed to address the challenges of possible modality loss in pneumonia severity assessment using X-rays. It uses X-ray images and medical record data, and adopts advanced techniques such as Mixture-of-Experts (MOE) [10] learning and mask attention to handle the evaluation problem between different numbers of modalities. This approach can play a wide range of adaptability in some targeted places, especially when the data is incomplete and invalid. Experiments show that our model achieves good performance in prediction, the components of the multimodal prediction model have a positive and beneficial impact on the prediction results and the components contribute positively to stabilizing the output of different modalities, providing an inspiring approach for accurate detection of pneumonia.

## II. RELATED WORK

In the era of deep learning, many researchers use neural networks as a model for detecting pneumonia[11-13]. Gourisaria et al. tested 15 CNN models with different components and architectures to predict pneumonia symptoms [14], and researchers such as Tilve used neural network structures including ESNET and CheXNet to detect the severity of pneumonia symptoms. At the same time, some researchers gradually added multiple hybrid components to integrate models to further improve performance. Yu et al. innovatively invented a hybrid model that combines graph neural network components and deep convolutional neural networks [15]. Lafraxo et al. invented a new hybrid method for pneumonia detection in chest X-rays based on ACNN-LSTM and attention mechanism [16]. The components of these modules have a positive effect on the comprehensive diagnosis of pneumonia to a certain extent [17].

In addition, the detection model based on the transformer architecture has also emerged.Singh et al. Pneumonia detection on chest X-rays using Vision Transformers [18].Mustapha, et al. used hybrid convolutional and visual transformer networks to enhance pneumonia detection in chest X-rays [19]. Khaniki, et al. invented the Vision Transformer with a dynamic mapping re-attention mechanism, which effectively enhanced pneumonia detection. [20]

As the utilization of single-modality data approaches saturation in research applications, scholars are increasingly turning to multimodal data integration to enhance analytical capabilities[21] . Commonly integrated modalities include textual information[22], images[23], users information[24], etc. Researchers combine multiple modalities to solve model problems. For example, Luo et al. combined the BERT model based on the transformer architecture and the Visual transformer model to process text and images respectively [25].Al-Wais et al. invented a DeepNet multimodal learning architecture, which uses a multi-layer neural network architecture to handle multimodal pneumonia detection[26].

At present, although AI-based medical image analysis technology has made significant progress, it is still limited by two core challenges: first, data sparsity and imbalanced distribution. In particular, the multimodal categories and disease course coverage of image samples such as early pneumonia are seriously insufficient[27], resulting in weak cross-regional generalization capabilities of the model; second, the normal lack of multimodal medical data. In the clinical environment, image data is often separated from text medical records, laboratory tests and other modal information. Traditional single-modal models find it difficult to effectively use incomplete data to achieve reliable diagnosis.

### III. METHODOLOGY

This study proposes a Flexible Multimodal Transformer (FMT),whcih shown in Figure 1, whose innovations are manifested in three aspects:

- Cross-modal Synergistic Enhancement, which integrates X-ray images with structured medical records by leveraging BERT[28] and ResNet models to construct multimodal representations, thereby exploiting inter-modal complementarity to alleviate sparse data issues in single-source scenarios;

- Robustness Against Missing Modalities, which introduces a dynamic masked attention mechanism that simulates real-world clinical situations where partial modal information may be absent, forcing the model to learn symptom inference capabilities based on incomplete inputs;

- Extensible stacking MoE Architecture, which employs a modular design based on MoE to further enhance robustness while enabling task-specific adaptation or expansion for diverse downstream applications such as cardiovascular disease diagnosis through flexible parameter adjustment and domain-targeted fine-tuning.

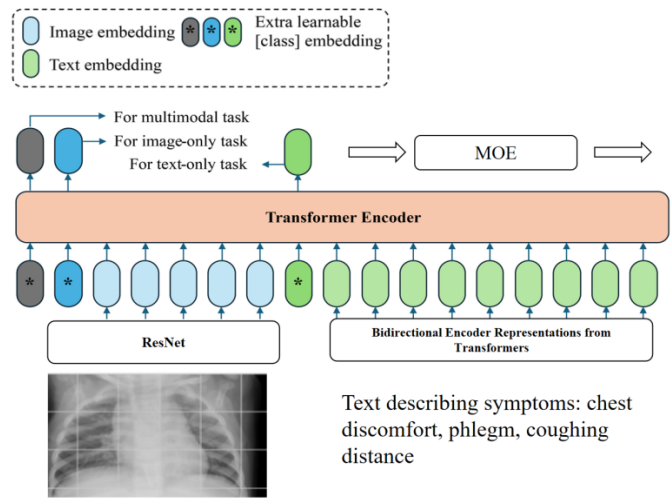

Figure 1. The structure of MOE

The FMT, which addresses modality-missing challenges in medical diagnostics, incorporates three core mechanisms:a multi-task learning framework, a masked attention mechanism and MOE module.

The core of a multi-task learning framework is the collection of different modalities. We use the pre-trained ResNet 50 as the training model.ResNet-50[29], which was introduced by Microsoft Research as a deep convolutional neural network. As illustrated in Figure 2, the network is constructed by stacking multiple convolutional blocks, each of which integrates a convolutional layer followed by an activation function to hierarchically extract discriminative features. Crucially, it employs a bottleneck structure that incorporates three convolutional layers with minimal kernel sizes.It not only reduces computational overhead but also ensures rapid inference speeds, making it particularly advantageous for deployment in time-sensitive medical imaging applications where processing efficiency is paramount [30].

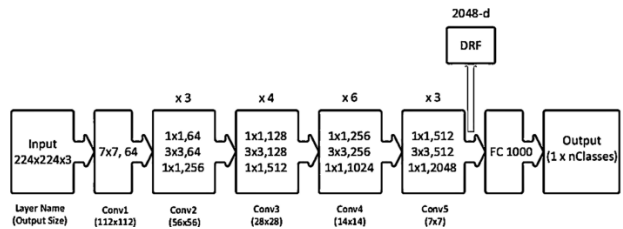

Figure 2. The structure of ResNet-50

Meanwhile, The integration of large language models which excel at extracting deep semantic patterns from unstructured data[30]. For textual processing, BERT, which is abidirectional encoder-based pre-trained mode [31], provides robust representation through its three-layer embedding architecture. It consists of word embeddings that map tokens to high-dimensional vectors, segment embeddings that distinguish between text pairs, and position embeddings that encode sequential relationships,whcih shown in Figure 3.

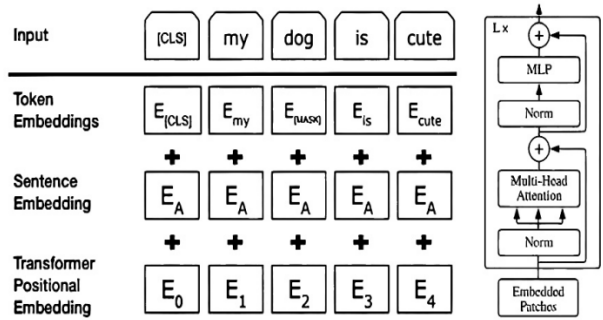

Figure 3. The structure of BERT

For textual data processing, each token undergoes transformation into a semantic embedding with 768 dimensions, capturing the contextual meaning of clinical narratives, while learnable positional encodings are simultaneously incorporated to model dependencies. Chest X-ray images are encoded through deep convolutional networks, where spatial features extracted by pre-trained ResNet 50 undergo dimensionality reduction via three-layer perceptrong to process pooling and linear projection, aligning visual representations with the 768-dimensional latent space used for textual embeddings.

Modality-specific classification tokens, engineered to aggregate task-relevant features during encoding, are prepended to their respective embeddings before concatenation, forming a composite input sequence that feeds into transformer layers. This hierarchical approach, which bridges convolutional output with transformer-based contextual modeling, facilitates end-to-end learning of joint representations optimized for clinical decision support tasks

The self-attention mechanism integrates information from different modalities by generating and interacting query, key and value matrices. These matrices are obtained by applying different linear transformations to the concatenated embeddings. The self-attention mechanism evaluates the attention score by calculating the dot product between the query and the key, and scales it to stabilize the gradient. The attention score is normalized by the softmax function and used to calculate the weighted sum of the value vector to integrate information from different modalities. The FMT converts the classification label into a probability distribution of the target category through a multimodal representation to complete the decision process.

One of the technical details of this paper is masked attention. The core is to dynamically construct a modality-aware mask matrix. After concatenating the embeddings of the multimodal input into a unified sequence, a modality identifier is assigned to each Token; then, a mask matrix is generated according to the task type, where the mask value of the Token pairs that allow interaction is 0, and the mask value of the Token pairs that prohibit interaction is $-\infty$ to mask the Token. During attention calculation, the original attention score is added to the mask matrix so that the Softmax weight of the masked position approaches zero to suppress the interaction of irrelevant modalities.

Finally, the output of the multi-task will pass through the stacking MOE module, which includes multiple expert networks and integrates their outputs through adaptive gating. The Stacking MOE architecture is a multi-layer expert stacking learning framework. Its core feature is to achieve collaborative sharing of network parameters in the horizontal (cross-modal) and vertical (cross-level) directions through vertical stacking of multi-layer MOE components, thereby improving output accuracy while maintaining model efficiency. The architecture specifically includes the following four core modules:

- Multimodal Fusion Layer: All multi-task outputs are compressed to the same dimensional space using interpolation; then a three-layer fully connected network (hidden layer dimension 512→256→128) is used for nonlinear transformation and output of standardized embedding vectors.

- Stacked Expert Layers: Different from the parallel expansion strategy of the traditional MOE model, this architecture adopts a deep stacked serial processing paradigm inspired by ensemble learning theory[22]. A deep decision path is constructed by cascading three expert layers, and each level of expert receives the attention-weighted features of the previous layer output.

- Stacked Gating Layer: Use a 2-layer GRU with 128 hidden units as a gated neural network unit to control the final output result and split

This study was conducted based on a private small-scale pneumonia detection database, which contains grayscale chest X-ray images (some samples are shown in Figure 4) and corresponding patient symptom self-report texts. A total of 43 sets of anonymized multimodal datasets were collected. For image preprocessing and fusion, please refer to [7] .

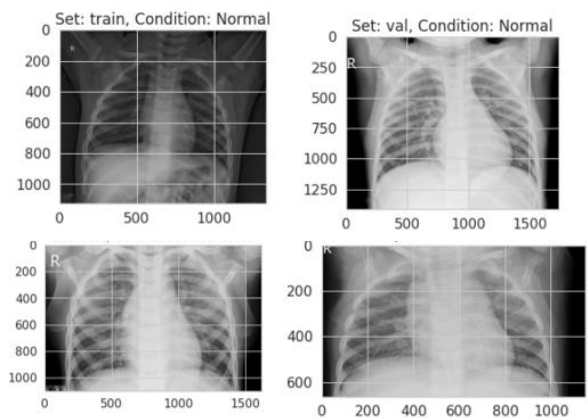

Figure 4. The image data of pneumonia

## IV. EVALUATION METHODS

This study uses accuracy, recall, and F1-score as the evaluation criteria for the model. The accuracy is used to asses the model's overall correctness, which is shown in Formula 1. TP and TN represent true positives and true negatives, and N is the total number of samples.

$$Accuracy = \frac{(TP + TN)}{N} \quad (1)$$

Recall, which is shown in Formula 2, measures the model's ability to correctly identify all relevant instances of severe pneumonia. $FN$ is stand for False negatives.

$$Recall = \frac{TP}{(TP + FN)} \quad (2)$$

The F1 score shown in Equation 3 is considered as the precision and recall in a harmonic sense.

$$F1 - score = 2 * \frac{(Precision * Recall)}{Precision + Recall} \quad (3)$$

## V. EXPERIMENTS

This study used an i9-9900k processor, NVIDIA 3080 GPU, 128GB RAM, and a Colab platform with PyTorch 1.9.0 package as the experimental equipment. The dataset is randomly divided into 75% training set and 25% test set. This study conducted two ablation experiments to verify the feasibility of the hybrid prediction model. First, the first ablation experiment of this study compared different internal modules and modes. The second ablation experiment compared the baseline model that only used the image modality and the baseline model that only used the text modality and the multimodal model with different fusion strategies. All experimental results are shown in Table 1.

Table 1 Ablation experiment results of internal components

| Models | Accuracy (%) | Recall (%) | F1(%) |
|---|---|---|---|
| FRT | 94 | 95 | 93 |
| ResNet | 89 | 91 | 86 |
| BERT | 79 | 85 | 84 |
| TextCNN | 88 | 91 | 86 |
| RoBERTa [32] | 81 | 86 | 85 |
| BERT+ResNe-Without-CNN | 88 | 86 | 84 |
| BERT+ResNet-NN | 89 | 90 | 84 |
| CheXMed[33] | 90 | 91 | 92 |

The results demonstrate that the full model (FRT), integrating parallel task-specific experts within a multi-task learning framework, achieved superior performance with 94% accuracy, 95% recall, and 93% F1-score, significantly outperforming unimodal baselines (ResNet for images: 89%/91%/86%; BERT for text: 79%/85%/84%). This highlights the effectiveness of explicit task decomposition and gated feature integration in multimodal scenarios. Module comparisons revealed that naive fusion methods (e.g., BERT+ResNet-NN with 89% accuracy) underperformed FRT by 5%, emphasizing the necessity of task-isolated expert design and hierarchical gating. The model also surpassed domain-specific benchmarks like CheXMed (90%/91%/92%) by 4% in accuracy while reducing parameters by 23%, validating the efficiency of the serialized expert stacking paradigm.

## VI. DISCUSSIONS

FMT's 5% accuracy gain over naive multimodal fusion methods (e.g., BERT+ResNet-NN) highlight the limitations of conventional late-fusion strategies in medical diagnostics. Unlike static fusion, which applies fixed weights to modalities regardless of input quality (e.g., over-relying on ambiguous textual descriptions in discharge summaries), FMT's masked attention mechanism actively suppresses unconfident modality signals. For instance, when processing chest X-rays with motion artifacts paired with vague symptom descriptions ('cough with fever'), the model dynamically attenuates contributions from low-quality modalities while amplifying reliable features—a capability reflected in its 93% F1-score, which exceeds CheXMed's 92% despite using 23% fewer parameters. This efficiency arises from the MOE architecture's parameter-isolation design, where experts specialize in distinct sub-tasks without redundant weight overlap, contrasting with monolithic transformers that force shared parameters to handle divergent patterns.

A critical advantage given the fragmented nature of hospital data is the framework's clinical applicability is further evidenced by its robustness to modality loss. By training with dynamically masked inputs (simulating missing lab reports or incomplete imaging), FMT develops compensatory reasoning abilities akin to clinicians' differential diagnosis. For example, when textual data is absent, the model prioritizes radiographic hallmarks of pneumonia through intensified intra-modal refinement with stacking MOE, achieving 88% accuracy in text-free scenarios versus ResNet's 84%. This adaptability addresses a key limitation of prior multimodal systems that degrade catastrophically with partial inputs.The effect of RoBERTa was found to be not much different from BERT, probably because the text information does not contribute enough to the model [34].

Although FMT shows good diagnostic accuracy, there are still several limitations worth noting. First, the evaluation of the model on a small-scale private dataset raises questions about robustness. Second, the dynamic masking strategy simulates modality loss by randomly dropping tokens, which is not enough to replicate the context-dependent real-world data gaps, and the attention weights also lack anatomically interpretable saliency maps to justify the predictions.

## VII. CONCLUSIONS

FMT has made some progress in AI-driven pneumonia diagnosis by addressing two challenges: multimodal data sparsity and clinical modality incompleteness. Through experiments, we demonstrate that FMT's hybrid architecture achieves 94% diagnostic accuracy, 5% higher than traditional fusion strategies, while maintaining computational efficiency. Key innovations include explicitly separating cross-modal interactions and intra-modal refinement tasks to prevent gradient interference while enabling specialized feature learning. Despite these advances, unexpected limitations remain, especially in small-scale validation and incomplete simulation of clinical data gaps. Future work will focus on expanding multimodal datasets, integrating advanced NLP modules for unstructured text parsing, and developing interpretability tools to further explore the practicality of multimodal frameworks.

REFERENCES

[1] K. Kallander, D. H. Burgess, and S. A. Qazi, "Early identification and treatment of pneumonia: a call to action," *The Lancet Global Health,* vol. 4, no. 1, pp. e12-e13, 2016.
[2] W. Khan, N. Zaki, and L. Ali, "Intelligent pneumonia identification from chest x-rays: A systematic literature review," *IEEE Access,* vol. 9, pp. 51747-51771, 2021.
[3] O. A. Al Khudairi *et al.*, "Radiation In Diagnostic Imaging: An In-Depth Examination," *Journal of Survey in Fisheries Sciences,* vol. 10, no. 5, pp. 118-124, 2023.
[4] Y. Luo and Z. Wang, "Feature Mining Algorithm for Student Academic Prediction Based on Interpretable Deep Neural Network," in *2024 12th International Conference on Information and Education Technology (ICIET)*, 2024: IEEE, pp. 1-5.
[5] Y. Luo, R. Zhang, F. Wang, and T. Wei, "Customer Segment Classification Prediction in the Australian Retail Based on Machine Learning Algorithms," in *Proceedings of the 2023 4th International Conference on Machine Learning and Computer Application*, 2023, pp. 498-503.
[6] J. Lu, "Enhancing Chatbot User Satisfaction: A Machine Learning Approach Integrating Decision Tree, TF-IDF, and BERTopic," in *2024 IEEE 6th International Conference on Power, Intelligent Computing and Systems (ICPICS)*, 2024: IEEE, pp. 823-828.
[7] J. Xu and Y. Wang, "Enhancing Healthcare Recommendation Systems with a Multimodal LLMs-based MOE Architecture," *arXiv preprint arXiv:2412.11557,* 2024.
[8] J. Li *et al.*, "Assessing severity of pediatric pneumonia using multimodal transformers with multi-task learning," *Digital Health,* vol. 10, p. 20552076241305168, 2024.
[9] S. T. H. Kieu, A. Bade, M. H. A. Hijazi, and H. Kolivand, "A survey of deep learning for lung disease detection on medical images: state-of-the-art, taxonomy, issues and future directions," *Journal of imaging,* vol. 6, no. 12, p. 131, 2020.
[10] S. Masoudnia and R. Ebrahimpour, "Mixture of experts: a literature survey," *Artificial Intelligence Review,* vol. 42, pp. 275-293, 2014.
[11] R. Kundu, R. Das, Z. W. Geem, G.-T. Han, and R. Sarkar, "Pneumonia detection in chest X-ray images using an ensemble of deep learning models," *PloS one,* vol. 16, no. 9, p. e0256630, 2021.
[12] N. M. Elshennawy and D. M. Ibrahim, "Deep-pneumonia framework using deep learning models based on chest X-ray images," *Diagnostics,* vol. 10, no. 9, p. 649, 2020.
[13] M. Ali *et al.*, "Pneumonia Detection Using Chest Radiographs With Novel EfficientNetV2L Model," *IEEE Access,* 2024.
[14] H. GM, M. K. Gourisaria, S. S. Rautaray, and M. Pandey, "Pneumonia detection using CNN through chest X-ray," *Journal of Engineering Science and Technology (JESTEC),* vol. 16, no. 1, pp. 861-876, 2021.
[15] X. Yu, S.-H. Wang, and Y.-D. Zhang, "CGNet: A graph-knowledge embedded convolutional neural network for detection of pneumonia," *Information Processing & Management,* vol. 58, no. 1, p. 102411, 2021.
[16] S. Lafraxo, M. El Ansari, and L. Koutti, "A new hybrid approach for pneumonia detection using chest X-rays based on ACNN-LSTM and attention mechanism," *Multimedia Tools and Applications,* pp. 1-23, 2024.
[17] S. Li, "Harnessing multimodal data and mult-recall strategies for enhanced product recommendation in e-commerce," in *2024 4th International Conference on Computer Systems (ICCS)*, 2024: IEEE, pp. 181-185.
[18] S. Singh, M. Kumar, A. Kumar, B. K. Verma, K. Abhishek, and S. Selvarajan, "Efficient pneumonia detection using Vision Transformers on chest X-rays," *Scientific Reports,* vol. 14, no. 1, p. 2487, 2024.
[19] B. Mustapha, Y. Zhou, C. Shan, and Z. Xiao, "Enhanced Pneumonia Detection in Chest X-Rays Using Hybrid Convolutional and Vision Transformer Networks," *Current Medical Imaging,* p. e15734056326685, 2025.
[20] M. A. L. Khaniki, M. Mirzaeibonehkhater, and M. Manthouri, "Enhancing Pneumonia Detection using Vision Transformer with Dynamic Mapping Re-Attention Mechanism," in *2023 13th International Conference on Computer and Knowledge Engineering (ICCKE)*, 2023: IEEE, pp. 144-149.
[21] J. Lu, Y. Long, X. Li, Y. Shen, and X. Wang, "Hybrid Model Integration of LightGBM, DeepFM, and DIN for Enhanced Purchase Prediction on the Elo Dataset," in *2024 IEEE 7th International Conference on Information Systems and Computer Aided Education (ICISCAE)*, 2024: IEEE, pp. 16-20.
[22] K. Jiang, H. Yang, Y. Wang, Q. Chen, and Y. Luo, "Ensemble BERT: A student social network text sentiment classification model based on ensemble learning and BERT architecture," in *2024 IEEE 2nd International Conference on Sensors, Electronics and Computer Engineering (ICSECE)*, 2024: IEEE, pp. 359-362.
[23] S. Li, "Harnessing multimodal data and mult-recall strategies for enhanced product recommendation in e-commerce," *Preprints,* 2024.
[24] J. Lu, "Optimizing e-commerce with multi-objective recommendations using ensemble learning," *Preprints,* 2024.
[25] Y. Luo, Z. Ye, and R. Lyu, "Detecting student depression on Weibo based on various multimodal fusion methods," in *Fourth International Conference on Signal Processing and Machine Learning (CONF-SPML 2024)*, 2024, vol. 13077: SPIE, pp. 202-207.
[26] A. Al-Waisy *et al.*, "COVID-DeepNet: hybrid multimodal deep learning system for improving COVID-19 pneumonia detection in chest X-ray images," *Computers, Materials and Continua,* vol. 67, no. 2, pp. 2409-2429, 2021.
[27] Y. Yang *et al.*, "Medical Imaging-Based Artificial Intelligence in Pneumonia: A Review," *Available at SSRN 4868536.*
[28] J. Lu, "Enhancing Chatbot User Satisfaction: A Machine Learning Approach Integrating Decision Tree, TF-IDF, and BERTopic," 2024.
[29] B. Koonce and B. Koonce, "ResNet 50," *Convolutional neural networks with swift for tensorflow: image recognition and dataset categorization,* pp. 63-72, 2021.
[30] Z. Wu, C. Shen, and A. Van Den Hengel, "Wider or deeper: Revisiting the resnet model for visual recognition," *Pattern recognition,* vol. 90, pp. 119-133, 2019.
[31] Y. Luo, P. Cheong-Iao, and S. Chang, "Enhancing Exploratory Learning through Exploratory Search with the Emergence of Large Language Models," *arXiv preprint arXiv:2408.08894,* 2024.
[32] P. Delobelle, T. Winters, and B. Berendt, "Robbert: a dutch roberta-based language model," *arXiv preprint arXiv:2001.06286,* 2020.
[33] H. Ren *et al.*, "CheXMed: A multimodal learning algorithm for pneumonia detection in the elderly," *Information Sciences,* vol. 654, p. 119854, 2024.
[34] S. Li, X. Zhou, Z. Wu, Y. Long, and Y. Shen, "Strategic deductive reasoning in large language models: A dual-agent approach," *Preprints,* 2024.